\documentclass[11pt,a4paper]{article}
\usepackage{geometry}
\usepackage[titletoc,title]{appendix}

\usepackage{graphicx}
\usepackage[dvipsnames]{xcolor}
\usepackage{color}
\usepackage{amssymb}
\usepackage{float}
\usepackage{makecell}
\usepackage{amsbsy}
\usepackage{flushend}
\usepackage{float}
\usepackage{amsmath}
\usepackage{adjustbox}
\usepackage{booktabs}
\usepackage{multirow}
\usepackage{wrapfig}
\usepackage{cite}
\usepackage{hyperref}
\hypersetup{
     bookmarks=true,         
     unicode=false,          
     pdftoolbar=true,        
     pdfmenubar=true,        
     pdffitwindow=false,     
     pdfstartview={FitH},    
     pdftitle={},    
     pdfauthor={},     
     pdfsubject={},   
     pdfcreator={},   
     pdfproducer={}, 
     pdfkeywords={}, 
     pdfnewwindow=true,      
     colorlinks=true,       
     linkcolor={blue},          
     citecolor=blue,        
     filecolor=blue,      
     urlcolor=blue           
}

\newcommand{\Real}{{\mathbb R}}

\newcommand{\I}{\mathbf{I}}
\newcommand{\bK}{\mathbf{K}}

\newcommand{\X}{\mathbf{X}}

\newcommand{\bk}{\mathbf{k}}

\newcommand{\x}{\mathbf{x}}
\newcommand{\y}{\mathbf{y}}

\newcommand{\dataset}{{\mathcal D}}
\newcommand{\Normal}[0]{\mathcal{N}}

\usepackage{fancyhdr}
\fancypagestyle{plain}{
  \fancyhf{}
  \fancyfoot[C]{\thepage}%
  \fancyhead[C]{\scriptsize \copyright IEEE. ACCEPTED FOR PUBLICATION IN IEEE TGRS. DOI 10.1109/TGRS.2017.2767205}
}
\fancypagestyle{firstpage}{
  \fancyhf{}
  \fancyhead[C]{\scriptsize \copyright IEEE. ACCEPTED FOR PUBLICATION IN IEEE TGRS. DOI 10.1109/TGRS.2017.2767205}
  \fancyfoot[L]{\scriptsize \copyright IEEE. Personal use of this material is permitted. Permission from IEEE must  be  obtained  for  all  other  users,  including  reprinting/republishing  this material  for  advertising  or  promotional  purposes,  creating  new  collective works for resale or redistribution to servers or lists, or reuse of any copyrighted components of this work in other works. DOI: 10.1109/TGRS.2017.2767205.\\
   This work was supported by the European Research Council (ERC) through the ERC-CoG-2014 SEDAL Project under Grant 647423. (Corresponding author: D. H. Svendsen, daniel.svendsen@uv.es).\\
	D. H. Svendsen, L. Martino, and G. Camps-Valls are with the Image Processing Laboratory . M. Campos-Taberner and F. J. Garc\'ia-Haro are with the Departament de F\'isica de la Terra i Termodin\'amica, Universitat de Val\`encia, 46010 Val\`encia, Spain.}
}

\begin{document}
\title{\textbf{Joint Gaussian Processes for\\
Biophysical Parameter Retrieval}} 
\date{}
\author{Daniel Heestermans Svendsen$^\ast$,~
        Luca~Martino$^\ast$,~
        Manuel Campos-Taberner$^\dagger$, \\Francisco Javier Garc\'ia-Haro$^\dagger$
        and~Gustau~Camps-Valls$^\ast$\\
        \textcolor{white}{space}\\
 		Universitat de Val\`encia\\
		{\footnotesize $^\ast$Image Processing Laboratory,
		$^\dagger$Department of Earth Physics and Thermodynamics}}



\maketitle

\begin{abstract}
Solving inverse problems is 
central to geosciences and remote sensing. Radiative transfer models (RTMs) represent mathematically the physical laws which govern the phenomena in remote sensing applications (forward models). The numerical inversion of the RTM equations is a challenging and computationally demanding problem, and for this reason, often the application of a nonlinear statistical regression is preferred. In general, regression models predict the biophysical parameter of interest from the corresponding received radiance. 
However, this approach does not employ the physical information encoded in the RTMs. An alternative strategy, which attempts to include the physical knowledge, consists in learning a regression model trained using data simulated by an RTM code. In this work, we introduce a nonlinear nonparametric regression model which combines the benefits of the two aforementioned approaches. The inversion is performed taking into account jointly both real observations and RTM-simulated data. The proposed Joint Gaussian Process (JGP) provides a solid framework for exploiting the regularities between the two types of data. The JGP automatically detects the relative quality of the simulated and real data, and combines them accordingly. This occurs by learning an additional hyper-parameter w.r.t. a standard GP model, \textcolor{black}{and fitting parameters through maximizing the pseudo-likelihood of the real observations}. The resulting scheme is both simple and robust, i.e., capable of adapting to different scenarios. The advantages of the JGP method compared to benchmark strategies are shown considering RTM-simulated and real observations in different experiments. Specifically, we consider leaf area index (LAI) retrieval from Landsat data combined with simulated data generated by the PROSAIL model.

\end{abstract}


%
\maketitle
\thispagestyle{firstpage}
\pagestyle{plain}

\section{Introduction}\label{Intro}

Solving {\it forward} and {\it inverse} problems lies at the heart of research in geoscience, remote sensing and physics in general. The {\em forward modelling} problem consists mainly in determining the physical laws which govern complex phenomena (for instance, modelling the response of a sensor to different physical inputs). Then, the designed model is implemented and tested in different scenarios in order to study the ability to explain observed physical phenomena. These forward mechanistic models are also used to generate artificial measurements~\cite{Snieder99}. In this work, we focus on  {\em radiative transfer models} (RTMs), which play the role of forward models in remote sensing applications of biophysical parameter estimation.

The aim of the {\em inverse} problem is to determine the underlying physical conditions which correspond to a given set of real obtained measurements. That is, it attempts to make inference about physical parameters from sensory data. A very relevant problem is that of estimating vegetation properties from remote sensing observations. Accurate inverse models help determine the phenological stage and health status (e.g., development, productivity, stress) of crops and forests~\cite{Hilker08}, which has important societal, environmental and economical implications, given the evergrowing demand for biofuel and food. Leaf chlorophyll content ($Chl$), leaf area index (LAI), biomass, and fractional vegetation cover (FVC) are among the most important vegetation parameters~\cite{Whittaker75,Lichtenthaler87}. 

Radiative transfer models (RTMs) are typically used to implement the forward direction~\cite{jacquemoud00,verhoef03}. However, inverting RTMs directly is very complex because the number of unknowns is generally larger than the number of independent radiometric information~\cite{Liang08}. Also, estimating physical parameters from RTMs is hampered by the presence of high levels of uncertainty and noise, primarily associated to atmospheric conditions and sensor calibration, sun angle, and viewing geometry, as well as the poor sampling of the parameter space in most of the applications. This translates into inverse problems where spectra deemed similar may correspond to diverse solutions. This gives rise to undetermination and ill-posed problems. 

Methods for solving the inverse problems (i.e., parameter retrieval) can be classified in three main families: {\it statistical}, {\it physical} (a.k.a., {\it numerical}) and {\it hybrid} inversion methods~\cite{CampsValls11mc}. The {\em statistical inversion} approach consists in applying a regression method in order to predict a bio-geo-physical parameter of interest such as LAI given observations obtained by the satellite. The regression models are trained using a collected dataset with data pairs formed by measurements obtained by the satellite (as input; e.g. reflectances) and from the corresponding parameter of interest (as output; e.g. LAI) measured {\it in situ}. Note that this approach is purely statistical since no physical information is employed for the parameter retrieval.

Perhaps the most widely used approach in remote sensing is the {\em physical or numerical inversion}, which uses the information provided by the physical laws in the parameter retrieval. Given a measured spectrum (e.g. from a satellite) and a suitable forward model, the idea is to compare with RTM-generated spectra in order to find the corresponding parameter of interest. This intutive approach for the inversion of RTMs is based on searching for similar spectra in a look-up-table (LUT), based on some similarity measure, and assigning the closest parameter~\cite{Verrelst2015273,CampsValls11mc}. Alternatively, more sophisticated strategies have been considered, for instance, the use of computational algorithms based on Bayesian schemes~\cite{ulrych2001bayes,CampsValls11mc}. Compared to statistical inversion, the physical inversion is more computationally costly in general, but it can yield more physically meaningful predictions for the parameters of interest. RTMs vary in complexity, based on the simplifications and assumptions made about the underlying physical phenomena. The cost of a more sophisticated physical model is computational complexity.

Finally, a last approach combines both statistical and physical inversion. The {\em hybrid inversion} applies a statistical regression model learning from artificial data generated by RTM simulations. That is, the hybrid inversion is similar to the statistical inversion but using simulated data only, instead of real data, for training the regression model~\cite{Verrelst20121832,Verrelst2015273,fang05,fang03,CampsValls16grsm}. The rationale behind this approach is clear: exploit the flexibility and speed of statistical learning algorithms trained on physically-meaningful data generated by an RTM.
The advantage w.r.t. the statistical inversion is that larger datasets can be used for training (instead of a few, possibly not representative, real in situ measurements) and the physical knowledge encoded in the RTMs is indirectly employed. However, the quality of the inversion depends again dramatically on the quality of the artificial data generated, i.e., the ability of the RTMs to mimic real data in different scenarios.

Hybrid inversion is very powerful and practical when no \emph{in situ} data are available. Indeed, hybrid inversion is currently an active field~\cite{Verrelst20121832,Verrelst12rse}, and is replacing physical inversion in many real applications and processing chains at local and global scale~\cite{Baret2007275,BusettoJSTARS}. However, it seems intuitive to let predictions be guided by actual measurements whenever they are present. The aim of this work is to combine the statistical and hybrid inversions keeping the benefits of both approaches. One trivial possible solution consists in  training the regression model considering a single dataset composed of the real and artificial data. However, when only very few real {\em in situ} measurements are available, the method can be very sensitive to the incorporation of simulated data from RTMs. The reason being that this naive approach does not take into account the differences between the statistical properties of the two types of data, and learns from both data sources without distinguishing them. As a consequence, the performance can be really poor and especially biased, depending on the quality of the RTM-simulated data. Another more sophisticated strategy consists in combining two different predictions obtained by independent regression models dedicated to each particular dataset (or piece of information), thus performing a sort of model combination~\cite{Wallis2011,Bordley1982,Scott2013,Luengo15icassp}, \cite[Chapter 8]{rasmussen06}. 
However, in this approach the different datasets are analyzed separately, hence the two regression models do not process all the available information, and may eventually lead to inconsistent (contradictory) predictions.

In this paper, we extend the hybrid inversion framework, proposing a statistical method which performs nonlinear and nonparametric inversion blending both real and simulated data with a suitable statistical approach. Our statistical model for parameter estimation is a Bayesian nonparametric approach known as Gaussian Processes (GPs)~\cite{rasmussen06}. GPs have yielded convincing results in recent years in many remote sensing and geoscience problems~\cite{Verrelst11gp,Verrelst12jstars,CampsVallsGRSL2013}. GPs provide state-of-the-art prediction accuracy results, confidence intervals for the predictions, and allow model specification and interpretation in solid probabilistic terms (for an up-to-date review of GPs in remote sensing, see~\cite{CampsValls16grsm}). The proposed method in this paper, called a {\em Joint Gaussian Process} (JGP) exploits the information contained in both datasets, and provides a solid framework for incorporating physical knowledge in GPs. It is particularly useful when the amount of in situ data is scarce and the simulated data are able to `fill in the gaps' of the input space, which incidentally is often the case in terrestrial campaigns. 
The JGP model is capable of automatically discovering the quality (noise, uncertainty) of each dataset, and including this information in the regression model to balance their trustworthiness.

The remainder of the paper is organized as follows. Section $\S2$ fixes notation, briefly reviews the GP framework and introduces the JGP. We illustrate performance of the JGP in a simple toy example, and comment on predictive mean in problems with multisource datasets. The JGP exploits the regularities between them, and provides a solid framework for incorporating physical knowledge in GP regression. Section $\S 3$ describes thoroughly the data used in the experiments. We rely on retrieval of LAI from Landsat observations and PROSAIL simulated data. Both the real in situ measurements and the simulations were targeted to rice crop monitoring in three top-producing areas in Europe, but the scheme and model is general enough to be extended to other cases. We give empirical evidence of performance in Section $\S 4$. We performed exhaustive experiments and comparisons in terms of accuracy and robustness, and discuss on the elusive concepts of hyperparameter tuning and extrapolation when uneven uncertainty levels and data scarcity are involved. We conclude in $\S 5$ with some remarks and an outline of future work.

\section{Joint Gaussian Processes}

Model inversion via regression is an old, largely studied problem in statistics and machine learning, as well as in remote sensing and geosciences. A large class of regression models are available in the literature, such as random forests, neural networks and kernel machines~\cite{Tramontana16bg,CampsValls11mc,CampsValls09wiley}. However, in the last decade, Gaussian processes have emerged as a solid framework to tackle prediction problems in general and in remote sensing in particular~\cite{CampsVallsGRSL2013,Verrelst20121832,CamposTaberner2015,CampsValls16grsm}. In this section, we first fix the notation, review the theory behind GPs, and propose the joint GP model.

\subsection{Gaussian Process (GP) regression}

Let us consider a dataset of $n$ pairs of measurements, ${\mathcal D}_n:=\{(\x_i,y_i)\}_{i=1}^n$. The input data pairs used to fit the inverse machine learning model $f(\cdot)$ might come from either {\em in situ} field campaign data (statistical approach) or simulations by means of an RTM (hybrid approach). Either way, let us assume a model of the form,
\begin{equation}\label{GLR}
	y_i = f(\x_i) + e_i,~e_i \sim\Normal(0,\sigma_e^2),
\end{equation}
where $f(\x)$ is an unknown latent function, $\x$ $\in$ $\Real^d$, and $\sigma_e^2$ is the noise variance. Now, if we define the vectors $\y$ $=$ $[y_1, \ldots ,y_n]^T$ and $\mathbf{f}$ $=$ $[f(\x_1),\ldots , f(\x_n)]^T$, the conditional distribution of $\y$ given $\mathbf{f}$ becomes $p(\y | \mathbf{f}) = \mathcal{N}(\mathbf{f}, \sigma_e^2\I)$, where $\I$ is the $n\times n$ identity matrix. At the heart of the GP approach, is the assumption that $\mathbf{f}$ follows a $n$-dimensional Gaussian distribution, in this case with zero-mean  $\mathbf{f} \sim \mathcal{N}(\boldsymbol{0}, \bK)$.
The covariance matrix ${\bf K}$, which defines the GP, is determined by a kernel function  $\bK_{ij}=k(\x_i,\x_j)$, encoding similarity between the input points~\cite{rasmussen06}. The intuition here is the following: the more similar input $i$ and $j$ are, according to some metric, the more correlated output values $i$ and $j$ ought to be. The most common kernel function to account for such similarity between points is the \emph{squared exponential} (SE) $k(\x_i,\x_j)=\exp(-\|\x_i-\x_j\|^2/(2\sigma^2))$, which has some advantages: it is a universal kernel function, it contains only one parameter that controls smoothness, and works in many  diverse areas of application.

It can be easily verified that the marginal distribution of $\y$ can be written as
\begin{equation}
	p(\y) = \int p(\y | \mathbf{f}) p(\mathbf{f}) d \mathbf{f} = \mathcal{N}(\mathbf{0},\boldsymbol{C}_n), \nonumber
\end{equation}
where $\boldsymbol{C}_n=\bK + \sigma_e^2\I $.
Now, what we are really interested in is regression, that is, in predicting a new output value $y_\ast$ given an input $\x_\ast$. The GP framework handles this by constructing a joint distribution over the training and test points,
\begin{equation}
	\begin{bmatrix}
		\y \\
		y_\ast
	\end{bmatrix}
	\sim
	\mathcal{N} \left( \mathbf{0},
	\begin{bmatrix}
		\boldsymbol{C}_n & \bk_\ast^T \\
		\bk_\ast & c_\ast
	\end{bmatrix} \right),  \nonumber
\end{equation}
where $\bk_{*} = [k(\x_*,\x_1), \ldots, k(\x_*,\x_n)]^T$ is an $n\times 1$ vector and $c_\ast = k(\x_*,\x_\ast) + \sigma_e^2$. Using standard manipulation of joint normally distributed variables~\cite{bishop2006pattern}, we can arrive at a distribution over $y_\ast$ conditioned on the training data. This is a normal distribution with predictive mean and variance given by
\begin{subequations} \label{eq:pred_dist}
\begin{align}
		\mu_{\text{GP}} (\x_\ast) &= \bk_{*}^T (\bK + \sigma_e^2\I_n)^{-1}\y, \label{eq:prediction} \\
		\sigma^2_{\text{GP}} (\x_\ast) &= c_\ast - \bk_{*}^T (\bK + \sigma_e^2\I_n)^{-1} \bk_{*}.
\end{align}
\end{subequations}
We see that GPs, apart from providing predictions $\mu_{\text{GP}\ast}$ for a given test input, also have a natural way of assessing the uncertainty of said predictions through the predictive variance (error bars) $\sigma^2_{\text{GP}\ast}$. The hyperparameters $\boldsymbol{\theta}=[\sigma, \sigma_e]$ to be tuned in the GP determine the width of the squared exponential kernel function and the model noise parameter, respectively. There are various ways to learn or infer the hyperparameters, including marginal log-likelihood maximization~\cite{rasmussen06}, simple grid search for least squares minimization or even recent combined strategies~\cite{Martino17eusipco}. In this work, we learn $\boldsymbol{\theta}$ using the so-called \emph{pseudo}-likelihood~\cite{rasmussen06}, the motivation and details of which will be explained below.

\subsection{Joint Gaussian Process (JGP) regression}\label{sec:jgp}

Let us now assume that the dataset $\dataset_n$ is formed by two disjoint sets: one set of $r$ real data pairs, $\dataset_r=\{(\x_i,y_i)\}_{i=1}^r$, and one set of $s$ RTM-simulated pairs $\dataset_s=\{(\x_i,y_i)\}_{i=r+1}^n$, so that $n=r+s$ and $\dataset_n=\dataset_r\cup \dataset_s$. In matrix form, we have $\X_r\in\Real^{r\times d}$, $\y_r\in \Real^{r\times 1}$, $\X_s\in\Real^{s\times d}$ and $\y_s\in \Real^{s\times 1}$,  containing all the inputs and outputs of $\dataset_r$ and $\dataset_s$, respectively. Finally, the $n\times 1$ vector $\y$ contains all the $n$ outputs, sorted with the real data first, followed by the simulated data. 

A naive approach to incorporating the information of the RTM would be to simply train a regular GP on the dataset $\dataset_n$, not allowing the model to differentiate between data-sources. This would accomplish our objective, that prediction ought to be guided by simulated data in the regions where real data is scarce\footnote{This is due to the covariance function defined by the squared exponential kernel, resulting in a high covariance of points that are close in input space.}. We know, however, that the distributions of the two datasets probably are not identical, and since we aim to predict points belonging to the {`real' distribution}, we suffer the problem that the auxiliary data might confuse our predictions in regions where we actually possess sufficient real data. 

In order to address this problem, a hyperparameter is added to the model, which controls how much the simulated data contributes to prediction. The altered covariance function takes the following form,
\begin{equation}\label{eq:jgpcov}
\boldsymbol{C}_n=\bK + \sigma_e^2\mathbf{V}, \,\,\, \mathbf{V} = \mbox{diag}(\, \underbrace{1,...,1}_r,\underbrace{\gamma^{-1},...,\gamma^{-1}}_s \,),
\end{equation}
where $\bK$ is now an $(r+s)\times(r+s)$ matrix, similar to the formulation of Bonilla et al. \cite{bonilla2008multi}, where we shall call $\gamma$ the \textit{trust parameter}. It has the straightforward interpretation, that it represents the modeled noise-variance in the simulated data, relative to that of the real data, e.g. a model of the form (cf. \eqref{GLR})
\begin{equation}
y_i = f(\x_i) + e_i,~ e_i \sim \Normal \left( \mathbf{0} , \begin{matrix}
      \sigma_e^2 ~ ~ ~ ~ ~ ~ \mbox{if} & i \leq r \\
      {\sigma_e^2}/{\gamma} ~ ~ ~ \mbox{if} & i > r
   \end{matrix} \right).
\end{equation}
We can also consider Eq.~\eqref{eq:prediction} written on kernel smoother form, $\mu_{\text{GP}} (\x_\ast) = \bk_{*}^T \boldsymbol{\alpha}$. A low trust parameter  quenches the components of $\boldsymbol{\alpha}$ pertaining to the simulated data points, thus damping their influence on prediction. We derive a discriminative alternative formulation to this probabilistic perspective of JGP in Appendix~\ref{app:lsjgp}, and a multisource formulation to deal with multiple datasets in Appendix~\ref{app:multisource}.

\subsection{Learning the hyperparameters}\label{sec:learning}
In this work, we want to make predictions with respect to the distribution of the real data so inferring hyperparameters must be done in accordance with this. The common scheme of marginal likelihood maximization~\cite{bishop2006pattern} is not effective because it attempts to maximize the likelihood of all data points simultaneously. We therefore propose to maximize the \textit{leave-one-out} (LOO) likelihood, also known as \textit{pseudo}-likelihood \cite{rasmussen06}, allowing us to maximize the likelihood of all data points but the simulated ones. This is reminiscent of the work by G. Leen et al. \cite{leen2012focused}, who construct a focused \textit{model}, where we in this work perform focused \textit{inference}.

The predictive probability of a single training data point conditioned on the remaining data is a normal distribution determined by Eq.~\eqref{eq:pred_dist}, using all data points but the $i$'th. Thus, the predictive log-likelihood leaving out training point $i$ can be expressed as
\begin{align*}
\log p(y_i | \X_{-i},\y_{-i}, \boldsymbol{\theta} ) =
-\frac{1}{2}\log 2\pi \sigma^2_i - \frac{(y_i - \mu_i )^2}{\sigma^2_i} .
\end{align*}
From this we can construct the LOO likelihood by summing over each data point and fit the hyperparameters to maximize it. We modify this approach here, by only summing over the real data points
\begin{equation}\label{LOO1}
	L_{\mbox{\footnotesize LOO}} (\X,\y, \boldsymbol{\theta}) = \sum_{i=1}^{r} \log p(y_i | \X_{-i},\y_{-i}, \boldsymbol{\theta}).
\end{equation}
In computing $r$ different predictive means and variances, it appears that we have to invert $r$ slightly different covariance matrices. Luckily, there is a way around this very computationally inefficient approach, which involves simply computing the inverse of the complete covariance matrix~\cite{sundararajanpseudo}. Instead of using Eq.~\eqref{eq:pred_dist} a total of $r$ times to evaluate the likelihood function, the following equations may be used:
\begin{equation}\label{LOO2}
\mu_i = y_i - \frac{[\bK^{-1}\y]_i}{\left[\bK^{-1}\right]_{ii}} \,\, , \,\,\,\, \sigma_i^2 = \frac{1}{[\bK^{-1}]_{ii}} \,\, ,
\end{equation}
where $[\,\cdot \,]_{i}$ denotes the $i$'th element of a vector, and $[\,\cdot \,]_{ii}$ is the $i$'th diagonal element of a matrix.

\subsection{Joint Gaussian Processes exemplified} \label{sec:jgpexample}

\begin{figure}[h!]
\begin{center}
\includegraphics[scale=0.45]{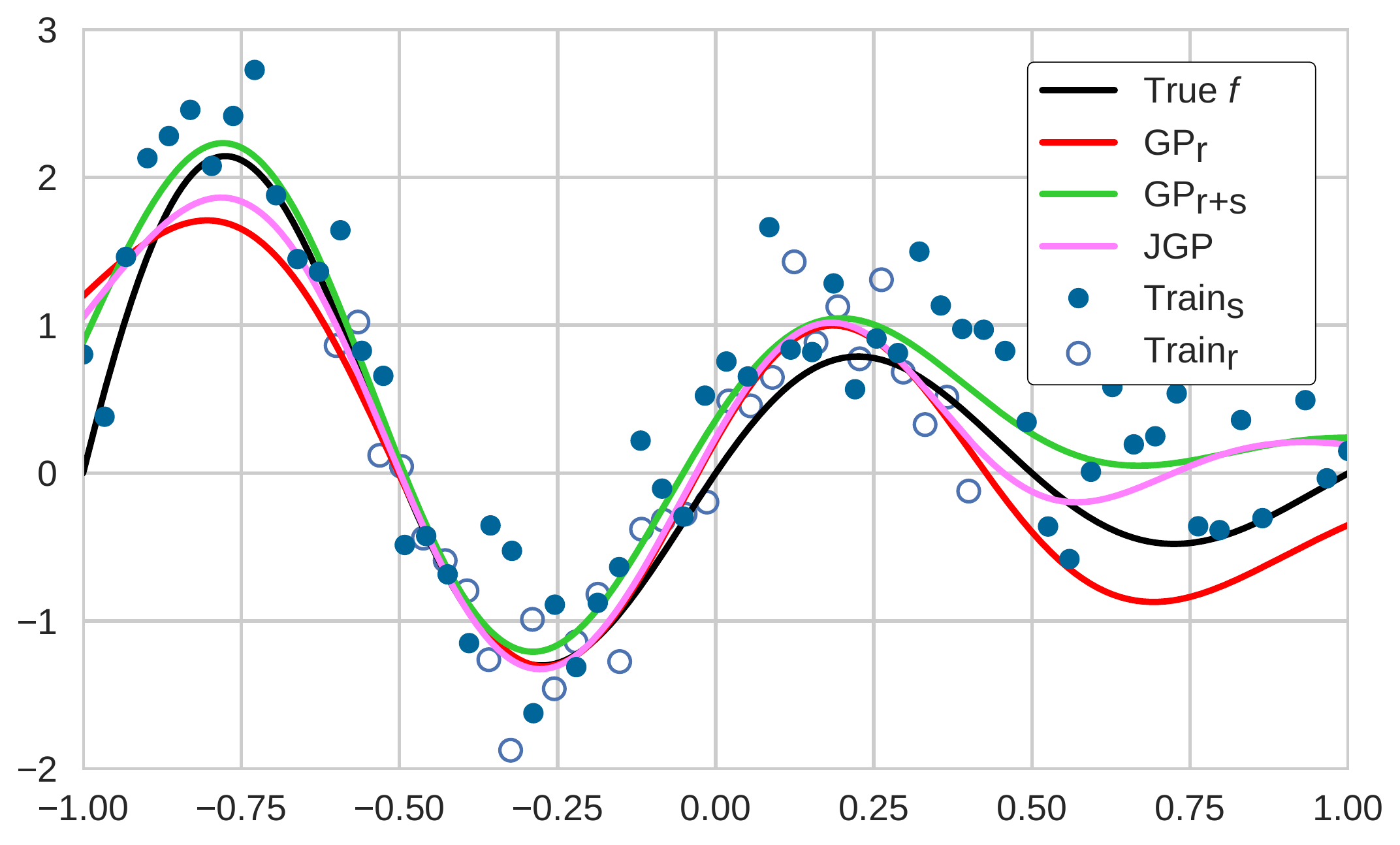}
\caption{Example of a Joint Gaussian Process in practice. \label{fig:illustrative}}
\end{center}
\end{figure}

Let us illustrate the solution of the JGP in a toy example. In Fig.~\ref{fig:illustrative} we include an illustrative example with real training points (subscript $r$) covering the range $[-0.6,+0.4]$, and simulated training points (subscript $s$) in the range $[-1,+1]$. Data were generated from the latent function in black, 
\begin{equation}\label{eq:modelito}
f(x) = b + \exp(-x) \sin(2\pi x) + \epsilon,
\end{equation} 
and buried in noise $\epsilon\sim{\mathcal N}(0,\sigma^2)$, where $\sigma=0.3$. We show the predictive mean of three GP models: One model trained on real data (in red), and one using real and simulated data together indiscriminitavely (in green). These models will be referred to as $\mbox{GP}_{r}$ and $\mbox{GP}_{r+s}$ respectively. Finally the JGP, also using both datatypes (in magenta). We assumed a SE covariance function and learned the optimal hyperparameters with the proposed LOO scheme. 

We observe three different regions in the figure. Below $x=-0.6$, we do not have real measurements, hence the GP$_r$ provides poor estimates, while both the GP$_{r+s}$ and the JGP model provide better fits to the generating function. At the center, $[-0.6,+0.4]$, we have a very accurate view of the latent function by all methods. For $x>0.4$, we do not have real training samples neither, so we observe the same behaviour as for low values: the GP$_r$ performs poorly revealing a strong bias, and the JGP model fits the observations better than GP$_{r+s}$ the latter does not weight the real data points sufficiently high in the overall solution. As commented before, the JGP can distinguish between real and simulated data, and weighs their information differently. This is especially convenient when predicting outside a data-rich, well-represented region, and can be intuitively seen as a `extrapolation' capability of the method.

\section{Data collection}
This section is devoted to describing the data used in the experiments. We describe the ground (in situ) dataset, the remote sensing images acquired over the study areas, and the simulations conducted using PROSAIL.

\subsection{Remote sensing and ground data}

\begin{figure*}[t!]
\begin{center}
\includegraphics[width=\textwidth]{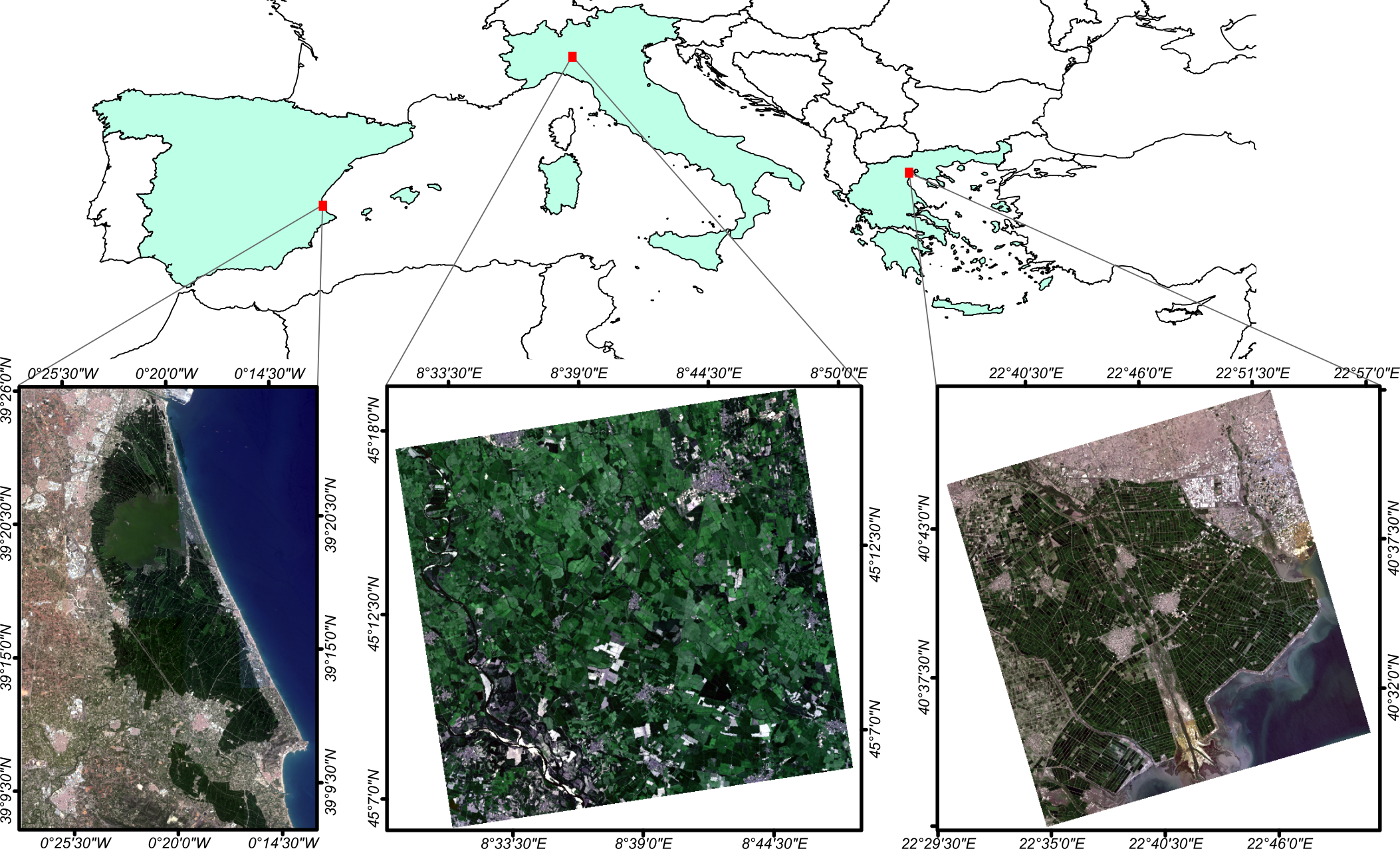}
\caption{\label{fig:study_areas}Study areas: Landsat 8 OLI surface reflectance RGB composites of the Spanish (left), Italian (middle) and Greek (right) study areas acquired on 3 August 2015, 18 July 2016 and  25 August 2016, respectively.}
\end{center}
\end{figure*}

The remote sensing and ground data used in this study were obtained in the framework of the ERMES project~\cite{BusettoJSTARS}. ERMES has developed an agro-monitoring system based on the assimilation of Earth observation and \emph{in situ} data for crop modelling solutions for rice monitoring. In this framework, non-destructive ground LAI data were acquired within rice fields in Spain, Italy and Greece (see Fig.~\ref{fig:study_areas}) during the 2015 and 2016 European rice seasons. The field campaigns were conducted from the very beginning of rice emergence (early-June) up to the maximum rice green LAI development (mid-August), and the temporal frequency of the measurements was approximately 10 days. This allowed for a multi-temporal database of \emph{in situ} LAI data covering the main phenological rice stages. The sampling was achieved selecting ESUs (Elementary Sampling Units) with different rice varieties and sowing dates in order to cover as much as possible the variability of the study areas, and the locations of the ESUs were far from the field borders. The same sampling scheme was adopted over each ESU, following the guidelines and recommendations of the VALERI (Validation of Land European Remote sensing Instruments) protocol. In addition, the centre of the ESU was geo-located to associate the mean LAI measurement with the corresponding satellite spectra.

LAI estimates were acquired in all three countries with smartphones using a dedicated smartphone app called PocketLAI~\cite{Confalonieri201367}, which was previously used in combination with GPs \cite{CamposTaberner2015}. PocketLAI uses both smartphone's accelerometer and camera to acquire images at 57.5$^\circ$ below the canopy and computes LAI through an internal segmentation algorithm~\cite{Confalonieri201367}. Specifically over rice fields, we have recently shown that LAI measurements taken with PocketLAI align well with other traditional acquisition instrumentation, such as plant canopy analyzers and digital cameras for hemispherical photography~\cite{CamposTaberner2015,CamposTaberner2016a}. A range of 18 to 24 measurements was taken over every ESU in order obtain a statistically significant mean LAI estimate per ESU.

Besides the aforementioned ground data, in this work we used Landsat-8 Operational Land Imager (OLI) and Landsat-7 Enhanced Thematic Mapper (ETM+) surface reflectance imagery. The images were downloaded through the United States Geological Survey (USGS), Earth Resources Observation and Science (EROS) and Center Science Processing Architecture (ESPA) during the 2015 and 2016 rice seasons over the three study areas. The provisional Landsat-8 Surface Reflectance (LaSRC)~\cite{Vermote2016} and the Landsat-7 ETM+ LEDAPS (Landsat Ecosystem Disturbance Adaptive Processing System) products (at 30 m spatial resolution) were used as inputs to retrieve Landsat-7/8 LAI estimates. The Landsat-7/8 surface reflectance spectral channels were filtered to relate only the blue (B), green (G), red (R), near infrared (NIR), and the two short wave infrared (SWIR1, SWIR2) bands with the ground LAI measurements in the retrieval process. Images were available every 16 day in Italy and Greece. On the other hand, since the Spanish rice area lies in two Landsat paths within the same row the temporal resolution of the images is increased up to seven and nine days.

\subsection{RTM Simulations}

In this paper, we simulated surface reflectance data of the selected study sites with the PROSAIL radiative transfer model. PROSAIL is the most widely used RTM in the last twenty years in remote sensing studies~\cite{jacquemoud09}. PROSAIL mimics canopy reflectance using the turbid medium assumption (i.e., assuming the canopy as a turbid medium for which leaves are randomly distributed), which is particularly well suited for homogeneous canopies like rice~\cite{Darvishzadeh2012,CamposTaberner2016b}. PROSAIL simulates leaf reflectance from 400 to 2500 nm with a 1 nm spectral resolution as a function of biochemistry and structure of the canopy, its leaves, the background soil reflectance and the sun-sensor geometry. Leaf optical properties are given by the mesophyll structural parameter (N), leaf chlorophyll (C$_{ab}$), dry matter (C$_{m}$), and water (C$_{w}$) contents. The water content was tied to the dry matter content ($C_{w}= C_{m}\times C_{wREL}/(1-C_{wREL})$) assuming that green leaves have a relative water content (C$_{wREL}$) varying within a relatively small range~\cite{Baret2007275}. At canopy level PROSAIL is characterized by the LAI, the average leaf angle inclination (ALA and the hot-spot parameter ($Hotspot$). In our experiments, the PROSAIL was run in forward mode for building a simulated data set (2000 pairs of Landsat-7/8 spectra and associated LAI) which was used for training purposes. In addition, a multiplicative brightness parameter ($\beta_{s}$) was applied to spectral rice background signatures (flooded and dry soil) to represent different background reflectance types~\cite{Baret2007275,Claverie2013216}. The system geometry was described by the solar zenith angle ($\theta_{s}$), view zenith angle ($\theta_{v}$), and the relative azimuth angle between both angles ($\Delta\Theta$). The distributions for the system geometry were randomly generated based on information in imagery metadata.

It is worth mentioning that in the case of simulating rice crops at high-resolution, sub-pixel non-vegetated areas located in the borders of rice fields, patches of bare/flooded soil, small water stripes and channels must be represented in the PROSAIL simulation~\cite{CamposTaberner2016b}. Hence, in order to account for these mixed conditions, we represented the pixels as a linear mixture of vegetation ($vCover$) and bare/flooded soil ($1-vCover$) spectra. A linear spectral mixing model was assumed for the sake of simplicity.

The leaf and canopy variables as well as the soil brightness and the $vCover$ parameter, were randomly generated following the parametrization in~\cite{CamposTaberner2016b,CamposTaberner2017} in order to constrain the behavior of the model to Mediterranean rice areas (see Table~\ref{tab:distributions}). In particular, a spectral library of underlying rice background (flooded and dry) signatures was used to obtain multitemporal LAI retrievals robust against changes in background condition related to water management.

\begin{table}[!h]
\begin{center}
\caption{Distribution of the canopy, leaf and soil parameters used in this work for simulation with the PROSAIL RTM.}
\vspace*{0.1cm}
\label{tab:distributions}
\renewcommand{\tabcolsep}{0.1cm}
\begin{tabular}{clccccc}
\toprule
\multicolumn{2}{c}{Parameter}                             & \multicolumn{1}{l}{Min} & \multicolumn{1}{l}{Max} & \multicolumn{1}{l}{Mode} & \multicolumn{1}{l}{Std} & Type         \\ \cmidrule(r){1-2} \cmidrule(r){3-7}
\multirow{4}{*}{Canopy} & LAI (m$^2$/m$^2$)                &0                         &10                         &3.5                          &4.5                         & Gaussian     \\
                        & ALA ($^\circ$)                          &30                         &80                         &60                          &20                         & Gaussian     \\
                        & Hotspot                          &0.1                         &0.5                         &0.2                          &0.2                         & Gaussian     \\
                        & vCover                          &0.5                        &1                         &1                          &0.2                        & Trunc. Gaussian     \\
\multirow{4}{*}{Leaf}   & N                                &1.2                         &2.2                         &1.5                          &0.3                         & Gaussian     \\
                        & C$_{ab}$ ($\mu$g$\cdot$cm$^{− 2}$) &20                         &90                         &45                          &30                         & Gaussian     \\
                        & C$_{dm}$ (g$\cdot$cm$^{− 2}$) &0.003                         &0.011                         &0.005                          &0.005                         & Gaussian     \\
                        & C$_{wREL}$                          &0.6                         &0.8                         &-                          &-                         & Uniform      \\
Soil                    & $\beta_{s}$                      &0.3   &1.2     &0.9     &0.25 & Gaussian \\ \bottomrule
\end{tabular}
\end{center}
\end{table}

\subsection{On the data distributions}

\begin{figure*}[t!]
    \begin{center}
	\includegraphics[width=\textwidth]{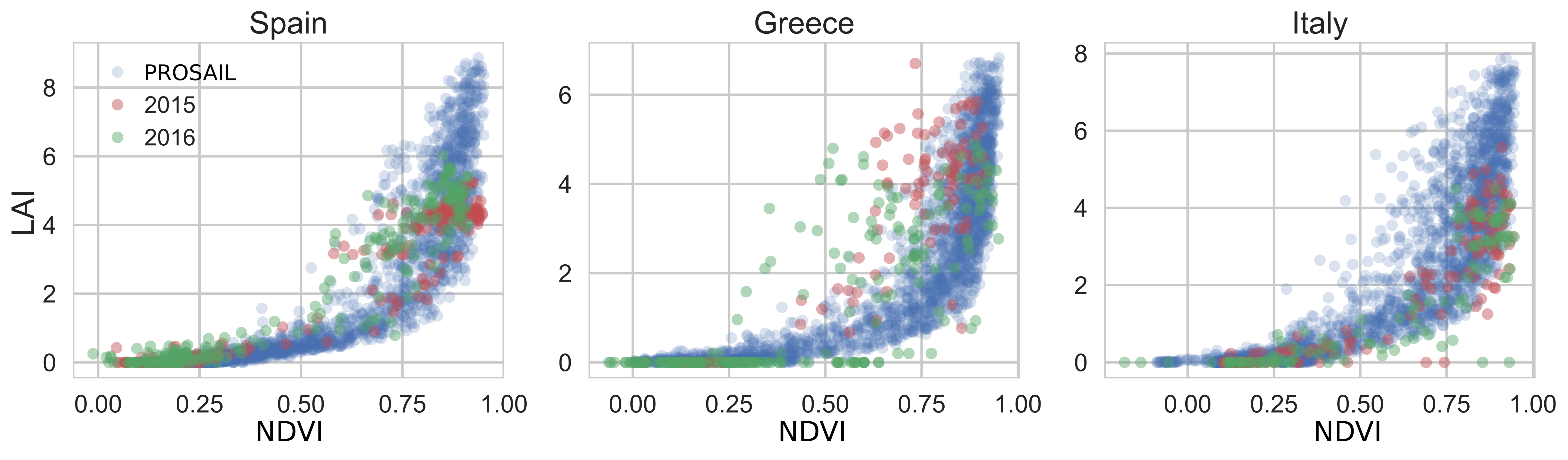}
    \end{center}
	\vspace{-0.5cm}
	\caption{Scatterplots in the NDVI-LAI representation space of the real and RTM-simulated data for all sites and acquisition campaigns (2015, 2016).}
	\label{fig:ndviscatter}
\end{figure*}

Blending in situ and simulated data requires a careful evaluation of the representativity of the data. When the distribution of the RTM-simulated data does not match the characteristics observed in real data, models using simulated data can be prone to error because learning good hyperparameters becomes a difficult task. Intuitively the JGP model tries to learn the relative relevance of both sources of information, which is impossible when datasets do not follow the same (or a similar) distribution. In this case, the model is most likely to disregard the information of the simulated data completely. It is important to remember that generating simulated data, through choosing sensible parameter ranges in PROSAIL is difficult, requires expert knowledge, and is scenario-dependent. Scatterplots in Fig.~\ref{fig:ndviscatter} show the distributions represented in the space of NDVI-vs-LAI for all sites and acquisition campaigns. These joint distributions suggest that the simulated points (in blue) cover regions of the greenness-LAI space efficiently for Spain and Italy, but cannot match the wide noise levels and variance observed in the real Greece data distributions, regardless the campaign. As we will see in the next experimental section, this has implications in the obtained results.

\section{Experimental results}
This section presents the experimental results obtained with the JGP model. We first evaluate the important issues of bias and noise variance in synthetic data distributions, and how the JGP deals with them. Then, empirical evidence of performance in two real experiments is given. First, we evaluate LAI prediction from Landsat images in all three sites and the two campaigns, and finally we analyze the performance in an extrapolation scenario.

\subsection{Robustness to bias and noise}

We use the same generating function as in section~\ref{sec:jgpexample} to illustrate the capabilities of the JGP to deal with systematic bias, and varying noise regimes (respectively, $b$ and $\sigma$ in Eq.~\eqref{eq:modelito}). In particular, we generated `real' datasets on the restricted interval shown in Fig. \ref{fig:illustrative}, and `simulated' datasets on the wider interval of $[-1;\,1]$ with different levels of white noise variance $\sigma_{sim}^2$ and values of an added bias $b_{sim}$. 
The test data was generated in the same way as the real training data, but over the entire interval $[-1;\,1]$, imitating a case of extrapolation where real data is unavailable in some domains, but simulated data of varying quality can be obtained across the whole representation space. 
We compare the performance of the JGP with the naive approach of training a regular GP on the combined datasets ($\mbox{GP}_{r+s}$), as well as that of a GP trained only on simulated ($\mbox{GP}_{s}$) or real data only ($\mbox{GP}_{r}$).  \textcolor{black}{The JGP hyperparameters are found by optimizing pseudo-likelihood over the real data, as described in section \ref{sec:learning}. The other methods also maximize pseudo-likelihood, however over all their data, for optimal comparison. These models might as well use standard marginal likelihood \cite{bishop2006pattern} maximization.}

\begin{figure}[H]
\begin{center}
\hspace*{-0.8cm}
\includegraphics[width=\textwidth]{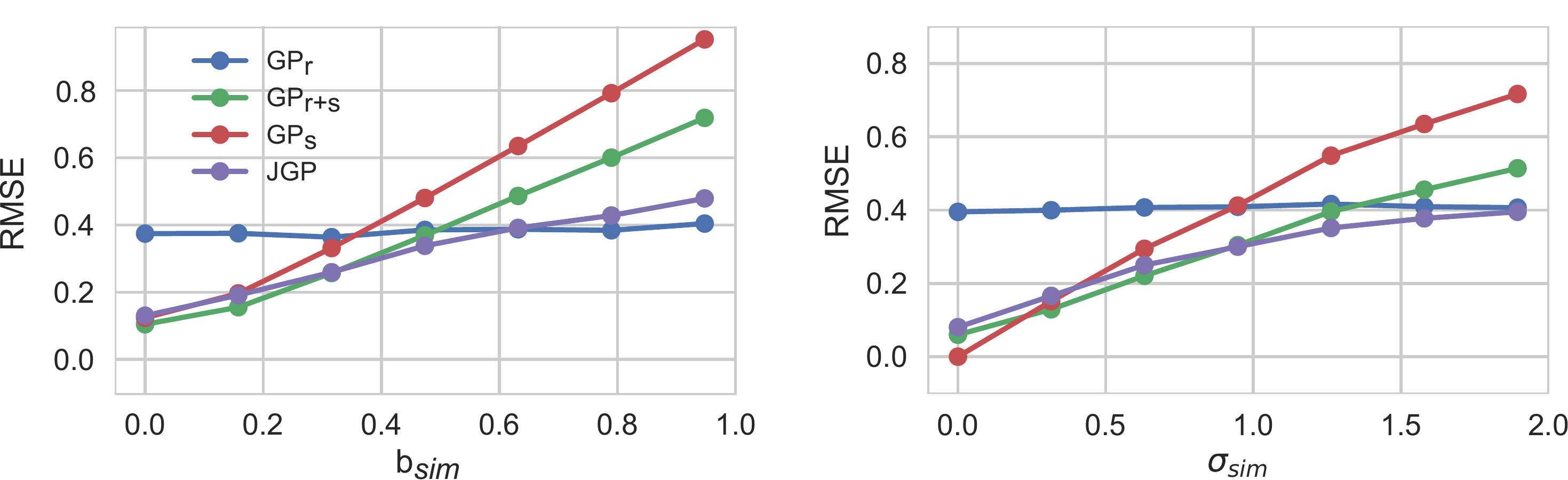}
\caption{\label{fig:toy} Performance of different schemes for including simulated data in a toy example where the quality of the secondary data source is varied.}
\end{center}
\end{figure}

From Fig. \ref{fig:toy} we can see the obvious result that if the simulated data is perfect, i.e. just points from the underlying damped sine, all methods that use the simulated data are performing better than the GP trained only on real data. Conversely, if the simulated data is very dissimilar the real, it is better not to use it at all. Depending on how the distribution of the two datasets diverge, there is a risk that the simulated data confuses the regression. 
We see that the JGP is the approach to incorporating simulated data which, roughly speaking, best handles a deterioration in the quality of these data points, be it through noisy regimes or systematic bias. 
The main takeaway here is that, as it is uncertain to know in advance how helpful and realistic simulations are, the JGP presents a safe way to incorporate physical knowledge about the inverse problem at hand. 

\subsection{LAI retrieval from Landsat images \label{sec:lailandsat}}

In this section, we assess the performance of the JGP and compare it to other ways of including simulated data when attempting to solve the inversion problem: The $\mbox{GP}_{r+s}$ and the $\mbox{GP}_{s}$ approach. To this end we use each of the six datasets collected through the campaigns in the respective countries (Spain, Greece, Italy) and years (2015, 2016). Root mean squared error (RMSE) is computed using a 10-fold cross-validation scheme. During each fold, the amount of simulated data used by the JGP and $\mbox{GP}_{r+s}$ is gradually increased in order to study how the ratio of simulated-to-real data points, call it $p$, affects method's performance. The full 2000 simulated points are used for the $\mbox{GP}_{s}$ since they are generated by PROSAIL to represent the target area as well as possible. Finally, the experiment is repeated 50 times to get stable results.

\begin{figure*}[t!]
    \begin{center}
	\includegraphics[width=\textwidth]{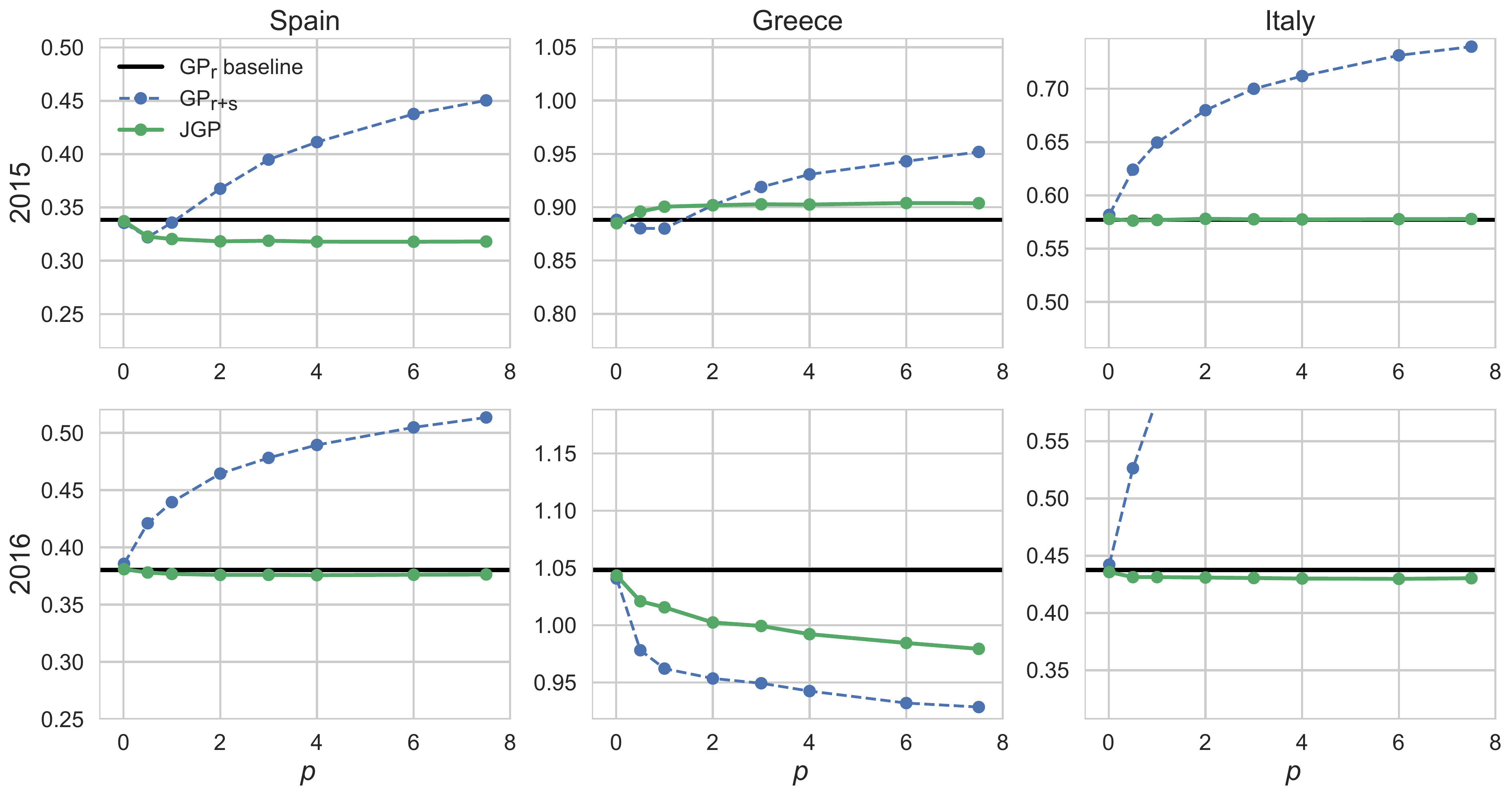}
    \end{center}
	\vspace{-0.5cm}
	\caption{Performance comparison (RMSE) for different ways of including simulated data. The JGP and the regular GP, trained on a dataset of real and simulated data pooled together, are compared to the base line of the GP trained exclusively on real data. RMSE is shown for the different sites, campaign dates and simulated-to-real data ratios. \textcolor{black}{As the scale is constant over the plots for better comparison, it was omitted from the plot in Italy 2016 how the $\mbox{GP}_{r+s}$ RMSE monotonically increases and reaches 0.85 for $p=8$.} }
	\label{fig:landsat}
\end{figure*}

The averaged RMSE as a function of the included simulated data is shown in Fig.~\ref{fig:landsat}. We observed rather different behaviours for the different datasets and scenarios. There are cases where $\gamma$ is fitted to a value close to 0, i.e. the JGP ignores the added data and simply follows the $\mbox{GP}_{s}$ baseline. For the datasets where this is not the case, we observe that relatively little simulated data is needed ($p\sim$\hspace*{0.05cm}$0.5$) to produce an effect. This is worth noting as the inversion of the kernel matrix, needed to train the JGP, scales in time complexity with the number of samples cubed, \textcolor{black}{ $\mathcal{O}(n^3) = \mathcal{O}((r+s)^3)$. This is equally true of the different methods used, regardless of the likelihood used. Furthermore, for predicting $m$ new points, we have $\mathcal{O}((r+s)m)$.}

In the case of Greece 2015, an average increase in RMSE is observed which, percentage-wise is around $\sim 1\,\%$. In Spain 2015 and Greece 2016, a decrease in RMSE of around $\sim 5\,\%$ can be observed. Interestingly, we see that the naive inclusion of simulated data (the $\mbox{GP}_{r+s}$ scheme) results in a general increase in error, except for the case of Greece 2016. This might be explained through the results shown in Fig. \ref{fig:toy} where $\mbox{GP}_{r+s}$ performs slightly better than the JGP approach when simulated data is of high quality. 

\begin{table*}[h!]
\small
\caption{Performance of $\mbox{GP}_{s}$ method \label{tab:simresults} \vspace*{-0.2cm}}
\begin{center}
\begin{tabular}{|c|c|c|c|}
\hline
RMSE $\mbox{GP}_{s}$ & Spain & Greece & Italy \\ \hline
2015 & 0.92 & 1.87 & 1.07\\ \hline
2016 & 1.09 & 1.32 & 1.20 \\ \hline
\end{tabular}
\end{center}
\end{table*}
\vspace*{0.5cm}

The approach of using only simulated data for predicting LAI, although it has been shown to aptly capture the temporal evolution of vegetation \cite{CamposTaberner2016b}, shows considerable predictive error, visually distorting the results of Fig. \ref{fig:landsat}. The performance of the $\mbox{GP}_{s}$ method is therefore instead given in Table \ref{tab:simresults}. Comparing with the baseline of Fig. \ref{fig:landsat}, we see that it suffers from a constant increase in RMSE of at least $25\%$. This underlines the point that, although RTM simulated data reflects the physical relation between input and output (spectrum and LAI), it struggles to mimic in situ data.\\

\subsection{LAI retrieval in extrapolation scenario}
\if
In this last experiment we aim to evaluate the performance of the JGP in the case of extrapolation. We simulate such scenario through generating a variety of datasets with high missing data points, which could be due to cloud coverage,  sampling in growing seasons only, erroneous in-situ measurements, etc. The goal is to show that in such scenarios, including simulated data on top of the scarce real data may help. 
\fi
In this last experiment, we demonstrate the `far extrapolation use-case' for the JGP method.
The experiment of section \ref{sec:lailandsat}, in practice, creates small `holes' in the real data distribution by removing a tenth for testing according to the $10$-fold cross-validation scheme. These holes in the representation space might then be filled with simulated data. The natural use however, is one where extrapolation is necessary into a region where no training data exists, but where a physical model might generate physically meaningful data points. 
Such scenarios might come about due to cloud coverage, unsystematic sampling in through growing seasons, erroneous in-situ measurements, etc.

\begin{table}[h!]
\small
\caption{RMSE of the $\mbox{GP}_{r}$, $\mbox{GP}_{r+s}$, $\mbox{GP}_{s}$ and JGP methods when dividing the real data so that test and training data are well-seperated domains. The top and the bottom rows (seperated by thick horizontal line) hold results from the 50-50 and 75-25 partition schemes respectively. \label{tab:extraresults}}
{\footnotesize
\begin{center}
\begin{tabular}{|r|c|c|c|}
\hline
{ \scriptsize $\mbox{GP}_{r}$/$\mbox{GP}_{r+s}$/$\mbox{GP}_{s}$/JGP} & Spain & Greece & Italy \\ \hline
\textbf{50-50} \hspace{1.1cm} 2015 \hspace{0.05cm} & 3.78/1.26/1.28/3.12 & 3.05/\text{1.30}/2.41/2.28 & 1.82/1.63 /\text{1.50}/1.56 \\ \hline
 2016 \hspace{0.05cm} & 4.31/1.65/\text{1.50}/2.92 & 2.90/\text{1.78}/1.87/2.69 & 1.91/1.36/1.70/\text{0.77} \\ \Xhline{6\arrayrulewidth}
\textbf{75-25} \hspace{1.1cm} 2015 \hspace{0.05cm} & 2.30/0.914/1.29/1.72 & 1.80/\text{1.29}/2.85/1.31 & 1.16/1.18/\text{1.77}/0.961 \\ \hline
 2016 \hspace{0.05cm} & 2.44/0.94/\text{1.59}/2.43 & 3.33/1.89/1.73/2.83 & 1.193/1.64 /2.22/\text{0.77} \\ \hline
\end{tabular}
\end{center}
}
\end{table}

A way to imitate such a scenario is to locate the green band median $\tilde{x}_{\textcolor{black}{G}}$ of the in-situ data and split it 50-50 such that the training and test data respectively have low and high values in the green band. This corresponds to the, rather unrealistic case, where sampling only takes place in the beginning of the year. \textcolor{black}{We also used the upper quartile of the green band to perform a 75-25 split of training-test data. A $p$ of 1 was chosen for the JGP and $\mbox{GP}_{r+s}$,  while the $\mbox{GP}_{s}$ was trained on all 2000 RTM-simulated datapoints as before.
The gain in performance in such a scenario is shown in Table \ref{tab:extraresults}, showing generally large RMSE reductions for all methods compared to $\mbox{GP}_{r}$, although the baseline is unreasonably high. In this experiment the JGP does the best when it during the training phase fits a high trust paramter, i.e. deems that the RTM-data is predictive of the real data. In Spain this does not appear to be occurring. It is however the only method which does not perform worse than $\mbox{GP}_{r}$ on any dataset.}

\section{Conclusion}

This paper introduced a method based on Gaussian Processes for biophysical parameter retrieval. To our knowledge, this is the first statistical non-parametric model blending in situ measurements and RTM-simulations. The model allows for the combination of {\em in situ} data and simulated data generated by an RTM code. The formulation of JGP only incorporates one additional trade-off hyperparameter that learn the relative importance of real and simulated data, and is related to the specific noise variance in each dataset. In the training-phase, pseudo-likelihood is maximized with respect to the real data only, which was shown to be a safe way of including simulated data. We studied the model in terms of accuracy, robustness to bias and noise regimes, and performed simulations in high missing data regimes. 

We illustrated the performance in the particular case of estimating LAI using Landsat images and simulated data from PROSAIL. Noticeable gains in accuracy were obtained in general. The model exploits the space coverage of RTMs in regions where real data scarcity hampers performance, while at the same time respecting the information provided by real data. Given the wide applicability of the JGP model, we foresee applications of the model in domains other than vegetation monitoring where few real data can be acquired yet a mechanistic model is available. \textcolor{black}{It is also worth noting that incorporation of RTM-simulated data is not restricted to Gaussian Processes, i.e. other regression methods could benefit from this as well.}

Future work is tied to study the capabilities of the model for transportability across space and time simultaneously. For that, we plan to incorporate anisotropic and invariant kernels. In this sense, manifold alignment could benefit the model, for example by projecting simulated data distributions into the real one before doing the regression. This in principle should reduce the problems of mismatching and representativity of the simulations. Finally, it is worth noticing that the JGP model is easily extended to deal with multi-set scenarios, as shown in Appendix~\ref{app:multisource}. Therefore, different campaigns, sites, and teams could receive different trust hyperparameters in the model. This actually relates to the field of multitask learning, which has received little attention in remote sensing data processing and for classification problems only.

\begin{appendices}
\section{Least squares JGP formulation}\label{app:lsjgp}

Let us derive, from a regularized kernel regression perspective, the JGP presented in Section~\S2. We will follow the same rationale as in standard least squares regression with kernel methods~\cite{CampsValls09wiley}. We are given input data matrices $\X_r\in\Real^{r\times d}$, $\X_s\in\Real^{s\times d}$, and the corresponding target vectors $\y_r$ and $\y_s$. We can define the collectively grouped data as $\X_n\in\Real^{n\times d}$ and $\y_n$. Let us now define two kernel feature mappings $\boldsymbol{\phi}_r, \boldsymbol{\phi}_s$ that map real and simulated data, respectively, to a Hilbert feature space, ${\mathcal H}$, which may be in principle of higher (possibly infinite) dimensionality than $d$, i.e. $H = dim({\mathcal H})\gg d$. We indicate the mapped data matrices as $\mathbf{\Phi}_r\in\Real^{r\times H}$ and $\mathbf{\Phi}_s\in\Real^{s\times H}$, respectively.

Now let us define the following cost function $\mathcal{L}$ that trades-off the prediction errors using real or simulated data, and the standard regularization parameter:
\begin{eqnarray*}\mathcal{L} = \|{\bf y}_r - \mathbf{\Phi}_r{\bf w}\|^2 + \lambda_1 \|{\bf y}_s - \mathbf{\Phi}_s{\bf w}\|^2 + \lambda_2 \|{\bf w}\|^2.
\end{eqnarray*}
Differentiating w.r.t. ${\bf w}$ and equating to zero, we obtain 
\begin{eqnarray*}
(\mathbf{\Phi}_r^{\top}\mathbf{\Phi}_r + \lambda_1 \mathbf{\Phi}_s^{\top}\mathbf{\Phi}_s + \lambda_2 \mathbf{I}) {\bf w} = \mathbf{\Phi}_r^{\top}{\bf y}_r + \lambda_1\mathbf{\Phi}_s^{\top}{\bf y}_s.
\end{eqnarray*}
Now, by applying the following representer's theorem~\cite{RieNag55}, ${\bf w} = [\mathbf{\Phi}_r^{\top} \,\, \mathbf{\Phi}_s^{\top} ]\boldsymbol{\alpha}=\mathbf{\Phi}_n\boldsymbol{\alpha}$, and multiplying from the left by $\mathbf{\Phi}_n$ yields the solution:
\begin{equation*}
\mathbf{\Phi}_n (\mathbf{\Phi}_r^{\top}\mathbf{\Phi}_r+ \lambda_1 \mathbf{\Phi}_s^{\top}\mathbf{\Phi}_s + \lambda_2 \mathbf{I}) \mathbf{\Phi}_n^{\top} \boldsymbol{\alpha} 
= \mathbf{\Phi}_n(\mathbf{\Phi}_r^{\top}{\bf y}_r + \lambda_1\mathbf{\Phi}_s^{\top}{\bf y}_s),
\end{equation*}
which can be expressed solely in terms of kernel matrices as
\begin{equation*}
(\bK_{nr}\bK_{rn} + \lambda_1\bK_{ns}\bK_{sn} + \lambda_2 \bK_{nn})  \boldsymbol{\alpha} = [\bK_{nr}\y_r \,\,\, \lambda_1\bK_{ns}\y_s],
\end{equation*}
and then the solution comes in closed-form as
$$
\boldsymbol{\alpha} = (\bK_{nr}\bK_{rn} + \lambda_2\bK_{ns}\bK_{sn}+\lambda_1\bK_{nn})^{-1}~[\bK_{nr}\y_r \,\,\, \lambda_1\bK_{ns}\y_s],
$$
where the subscripts of the kernel matrices, which come about from the interpretation that the kernel function defines the inner product on the space ${\mathcal H}$, indicate their sizes and the samples involved in their calculation. 
Note that when $\lambda_1=0$ the standard kernel ridge regression (or equivalently the predictive mean for the standard GP) is obtained, otherwise $\lambda_1$ acts as an extra regularization term accounting for the relative importance of the real and the simulated data points. 

By grouping terms, and defining the diagonal matrix $\mathbf{V} = \mbox{diag}(1,\ldots,1, \lambda_1,\ldots,\lambda_1)$, and by identifying the regularization term as the noise term in the probabilistic view of GPs (i.e.,  $\lambda_2 = \sigma_e^2$), we reach the equivalent JGP model in Eq.~\eqref{sec:jgp}, with the simple solution of the predictive mean as $\boldsymbol{\alpha} = (\mathbf{K}_{nn} +\sigma_e^2 \mathbf{V})^{-1}{\bf y}$. With a pure discriminative approach one misses the probabilistic interpretation of the model and hyperparameters, and restricts oneself to mean predictions only.

\section{Multisource JGP formulation}\label{app:multisource}

The JGP formulation as presented in the paper assumes access to two datasets only: One coming from a `main' distribution according to which we wish to make predictions (real in situ data in our experiments), and one coming from an  `auxilliary' distribution (simulated data from an physical RTM model in our case). We could also generalize the formulation and assume that we access $m$ such auxiliary datasets $\{\dataset_{1}, \, \dataset_{2}, \, ... \, , \dataset_m\}$ each holding a different number of data points $\{s_1, \, s_2, \, ... \, , s_m\}$. The JGP can easily be extended to this multisource scenario by fitting a trust parameter to each dataset. The $\mathbf{V}$ matrix of the covariance function in Eq. \eqref{eq:jgpcov} simply becomes
\begin{equation}
\mathbf{V} = \mbox{diag}(\, \underbrace{1,...,1}_r,\underbrace{\gamma_1^{-1},...,\gamma_1^{-1}}_{s_1} \, , ..... \, , \underbrace{\gamma_m^{-1},...,\gamma_m^{-1}}_{s_m} \,),
\end{equation}
and the same solution applies. Note the relation of this multisource JGP to multitask formulations previously presented in remote sensing data classification~\cite{Leiva2012}.

\end{appendices}

\section*{Acknowledgment}

The authors would like to thank the  Institute for Electromagnetic Sensing of the Environment (CNR-IREA), the Cereal Institute of DEMETER and the Aristotle University of Thessaloniki (AUTH) for providing the Italian and Greek field data acquired under the ERMES project.

\bibliographystyle{ieeetr}
\bibliography{bibMANU,DSPKM,erc,BOOKbib,gpc}

\end{document}